# Mixed Reality Depth Contour Occlusion Using Binocular Similarity Matching and Three-dimensional Contour Optimisation

Naye Ji[1,2]   Fan Zhang[1,2]*   Haoxiang Zhang[1,2,3]   Youbing Zhao[1,2]   Dingguo Yu[1,2]

**Abstract**
Mixed reality applications often require virtual objects that are partly occluded by real objects. However, previous research and commercial products have limitations in terms of performance and efficiency. To address these challenges, we propose a novel depth contour occlusion (DCO) algorithm. The proposed method is based on the sensitivity of contour occlusion and a binocular stereoscopic vision device. In this method, a depth contour map is combined with a sparse depth map obtained from a two-stage adaptive filter area stereo matching algorithm and the depth contour information of the objects extracted by a digital image stabilisation optical flow method. We also propose a quadratic optimisation model with three constraints to generate an accurate dense map of the depth contour for high-quality real-virtual occlusion. The whole process is accelerated by GPU. To evaluate the effectiveness of the algorithm, we demonstrate a time consumption statistical analysis for each stage of the DCO algorithm execution. To verify the reliability of the real-virtual occlusion effect, we conduct an experimental analysis on single-sided, enclosed, and complex occlusions; subsequently, we compare it with the occlusion method without quadratic optimisation. With our GPU implementation for real-time DCO, the evaluation indicates that applying the presented DCO algorithm can enhance the real-time performance and the visual quality of real-virtual occlusion.

**Keywords** mixed reality· real-virtual occlusion· binocular camera· depth contour extraction·occlusion handling

## 1 Introduction

As an extension of augmented reality (AR) and virtual reality (VR), mixed reality (MR) (Milgram et al. 1995) will play an increasingly important role in various fields in the future. Real-virtual occlusion is one of the basic requirements to break the boundary between virtuality and reality. In contrast, current MR technology solely superimposes virtual objects on the real scene applying a three-dimensional (3D) registration and tracking method. One of the major problems of real-virtual occlusion, which are yet to be overcome, is depth handling. With accurate and efficient depth handling technology, occlusion and accurate tracking registration between virtual and real objects can be realised in real time; namely, fusion consistency and geometric consistency.

Currently, an important factor that hinder marketisation of MR technology is that efficient real-virtual occlusion algorithms rely on professional device. For example, Microsoft's HoloLens optical perspective MR glasses are equipped with professional modules, such as RGB-depth (RGB-D) cameras and eye movement sensors, which

Naye Ji
e-mail: jinaye@cuz.edu.cn
✉Fan Zhang
e-mail: fanzhang@cuz.edu.cn
Haoxiang Zhang
e-mail: 21851051@zju.edu.cn

[1] Intelligent Media Institute, Communication University of Zhejiang, Hangzhou, China
[2] Key Lab of Film and TV Media Technology of Zhejiang Province, Hangzhou, China
[3] Software Engineering School, Zhejiang University, Hangzhou, China

considerably simplify the complexity of the depth algorithm and improve the real-time performance of real-virtual interaction algorithm. However, the high cost of professional devices prevents this product from being popular. Another adverse factor is that even efficient real-virtual occlusion algorithms that do not rely on professional devices cannot guarantee accuracy and efficiency. In terms of the examples of Google ARCore and Apple ARKit, ARKit exploits machine learning algorithms to segment the silhouette of each character in a video sequence frame and subsequently renders the background, characters, and virtual objects according to the depth information to position the character in front of the virtual objects. ARCore demonstrates its core functions from the key points of calibration, tracking and consistent illumination. Recently, ARCore can handle some real-virtual occlusion shipped through its depth API.

However, there still exists many real-virtual occlusion challenges to solve. In the future, real-virtual occlusion in MR will be satisfied as follows: (1) real time, because MR is a real-time interactive scene; (2) contour sensitivity—the occlusion contour accuracy of the occluded virtual object requires greater accuracy than the depth accuracy, because the depth information provides the object's distance from the viewpoint but a contour requires a close fit with the contour of real-world objects; (3) portability—the proposed real-virtual occlusion algorithm can be implemented in a variety of environments.

In this paper, we propose a novel depth contour occlusion (DCO) algorithm. Particularly, it is combined with the sparse depth map obtained from a two-stage adaptive filter area stereo matching algorithm and the 3D depth contour of the objects and a quadratic optimisation model with three constraints to generate an accurate depth map of the depth contour. Finally. We realised the DCO algorithm to solve the real-virtual occlusion problem in MR.

The technical contributions of this study can be summarised as follows.

(1) We propose a novel binocular stereo matching algorithm based on two-stage adaptive binocular camera to generate a sparse depth map.
(2) We combine a bidirectional optical flow field algorithm and a quadratic optimisation model to generate accurate depth contours for high-quality real-virtual occlusion.
(3) We improve the efficiency of the real-virtual occlusion contour with a GPU-based speeding-up.

This paper is organised as follows. Section 2 presents an overview of the related work. Section 3 provides an overview of the proposed method. Section 4 introduces a two-stage adaptive absolute difference (AD)-census stereo matching to generate a sparse depth map. Section 5 details the depth contour extraction based on a digital image stabilisation (DIS) optical flow method. For the optimisation of depth information extraction, Section 6 proposes three constraints to adjust the strength of the depth, smoothness, and stability constraints during the quadratic optimisation. Section 7 introduces the implementation of the DCO algorithm and demonstrates the evaluation of the effectiveness of the algorithm and the reliability of the real-virtual occlusion effect. Finally, Section 8 concludes this paper and provides the scope of future work.

## 2 Related work

We give a brief overview of the current treatment methods for the real-virtual occlusion in AR/MR, which are mainly divided into four categories, partly according to the survey paper (Xin and Peng 2018): (1) pre-modelling method, (2) 3D reconstruction method, (3) contour-based approaches, and (4) deep learning approaches.

2.1 Pre-modelling method

Breen et al. (1996) and Klein et al. (2004) proposed a method to realise real-virtual occlusion based on a pre-modelling method. They established a corresponding model based on real objects in advance to complete real-virtual interaction and achieved relatively accurate real-virtual occlusion and collision between virtual and real objects. Fischer et al. (2004) deployed medical device and improved the modelling method for medical volume datasets, which extracts their visual hull volume. The resultant visual hull iso-surface, which is simplified significantly, is implemented for real-time static occlusion handling in their AR system. Although these methods maintain high precision and satisfactory real-time performance, their method requires pre-modelling, which cannot meet the versatility requirements and does not apply to common complex environments.

2.2 Three-dimensional model reconstruction

Fuchs et al. (1994) were the first who attempted to solve the problem of real-virtual occlusion and dealt with the occlusion problem in the video perspective AR system. However, their method relied on large-scale data capture device; therefore, the 3D reconstruction speed and accuracy were too low to meet the requirements of real-time interaction. Wloka et al. (1995) presented a stereoscopic video image matching algorithm, which leverages the change of the vertical coordinate of each pixel to calculate the depth of field information for the occlusion processing of the video perspective AR system. However, owing to the insufficient development of the stereo matching algorithm and the inadequate hardware conditions at that time, the occlusion effect was not ideal and the algorithm did not meet the requirements of real-time processing (three frames per second).

Ni et al. (2006) combined the fast sum of absolute differences (SAD) algorithm to build a depth detection system. The SAD algorithm could achieve a real-time processing speed of 30 frames per second. However, problems still exist, such as the high matching error rate in low texture areas and the low contour accuracy of occlusion. Thereafter, the research on the occlusion based on 3D reconstruction has focused on how to improve the accuracy and the efficiency of occlusion, for example, by only performing partial 3D reconstruction of areas that may be occluded to improve the processing speed, and by using offline 3D reconstruction (Tian et al. 2015), preload static scenes, or semi-global matching (SGM) algorithm to improve the reconstruction accuracy (Guo et al. 2018).

The advantage of the 3D reconstruction algorithm is that it can deal with almost every real environment without collecting information from the real scene in advance (Zheng, 2016). In contrast, it can solve the origin of the real-virtual occlusion problem. However, the disadvantage of this type of algorithm is that it requires a significant amount of calculation, and its accuracy is not as good as that of the pre-modelling algorithm in some special situations, such as static scenes.

2.3 Contour-based approaches

Berger et al. (1997) were the first who proposed a method based on object contour recognition to determine the occlusion relationship and realised a method that can quickly calculate the occlusion mask without 3D reconstruction. Although this method significantly improves the efficiency and the accuracy of occlusion processing, the effect of the algorithm is heavily dependent on the quality of contour mapping. If the foreground and background colours are similar, the recognition error is large and the occlusion mask cannot be generated accurately.

Feng (2007) modelled the virtual scene hierarchically, realising multilevel real-virtual occlusion, including occlusion for nonrigid bodies. However, the occlusion effect of characters and the accuracy of occlusion contours need improvement; moreover, the occlusion is not flexible.

A depth-based approach by Schmidt et al. (2002) used a binocular stereo camera system for computing dense disparity maps to combine real and virtual worlds with proper occlusions. The method is based on area matching and facilitates an efficient strategy by using the concept of a three-dimensional similarity accumulator, whereby occlusions are detected and object boundaries were extracted correctly. However, the raw depth measurements suffer from holes, low resolution and significant noise around the boundaries.

Some studies focus on RGB-D camera to handle the occlusion (Du et al. 2016; Walton and Steed 2017; Hebborn et al. 2017; Jorge et al. 2019; Luo et al. 2019). The main idea is to snap depth edges towards their corresponding colour edges and exploit this information to enhance the depth map, which is later used for depth testing to achieve dynamic occlusion effects.

The advantages of this type of method are that it is more efficient than the 3D model reconstruction method and the contour extraction is more accurate. However, when processing complex occlusion, the comprehensive performance of this type of algorithm is not as good as that of the method described above. The method based on contour extraction can alleviate the problem of insufficient hardware conditions and the low level of 3D reconstruction accuracy. Furthermore, it can be developed into a general real-virtual occlusion algorithm in the long term.

2.4 Deep learning approaches

Deep learning strategies estimate depth from colour data, and then convert the monocular camera into an RGB-D sensor (Luo et al. 2020). Unfortunately, only a few monocular depth map estimation techniques could achieve a performance compatible with the AR applications (Macedo and Apolinario 2021). Nonetheless, some emerging deep learning technologies are related to AR, such as single-image occlusion estimation (Lu et al. 2019) and saliency determination (Borji 2019).

Tang et al. (2020) presented GrabAR, which enhanced AR applications with interactions between hand (real) and grabbable objects (virtual) by formulating a compact deep neural network that learned to generate the occlusion mask. However, to train the network, a large dataset was needed. In addition, the method only achieved the occlusion of hands. Our proposed solution could achieve the occlusion of more diverse objects.

## 3 Overview of the proposed method

An overview of our proposed method is shown in Fig. 1. Specifically, a two-stage adaptive filter area stereo matching algorithm, called AD-census stereo matching, is implemented to establish a sparse depth map. The DIS optical flow method combined with the Canny algorithm is used to extract the depth contour of the object. According to the principle of depth discontinuity at the 3D edge contour, a quadratic optimisation model with three constraints is established to propagate the sparse depth information to dense depth information. Finally, a virtual object is generated in the world coordinate system, and if the depth value of the virtual object`s pixel is greater than the depth value of the pixel corresponding to the image, it is discarded during the rendering of the virtual object pixels.

In addition, in the proposed method, the left and right images of the binocular fisheye camera have been calibrated and stereo corrected to achieve the epipolar constraint of binocular matching by computing the un-distortion and rectification transformation map with OpenCV built-in De-distortion functions in initialization phase and CUDA accelerating at image cropping and remapping stage of each frame.

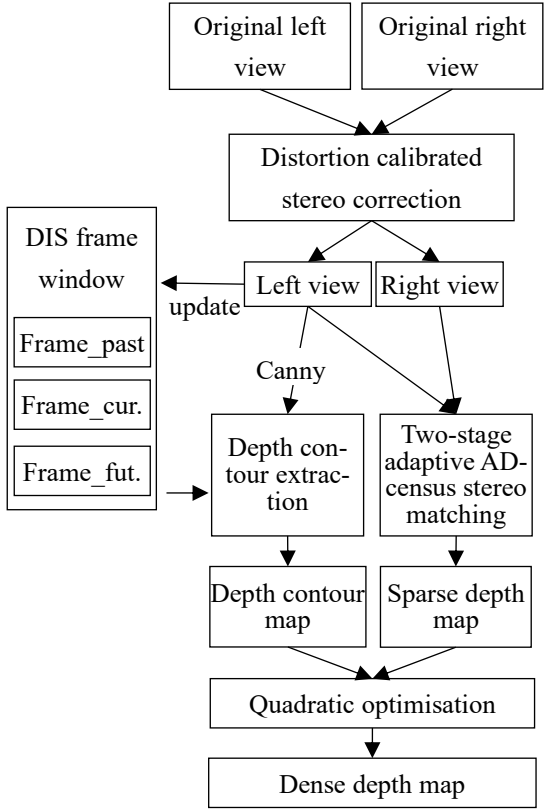

**Fig. 1** Overall architecture of the depth contour occlusion algorithm

## 4 Two-stage adaptive AD-census stereo matching

For real-virtual occlusion task, the depth information of the scene is required first. This paper deploys a two-stage adaptive AD-census stereo matching algorithm to generate sparse depth map. The algorithm is designed adaptively in both cost calculation and aggregation stage, which substantially improves the matching precision of the algorithm for weak texture and is suitable for most situations. In addition, to improve the stereo matching efficiency, we reduce the input image size to a quarter of its original size and finally discretised the sparse depth map into the original size.

4.1 Initialisation of the filtering area of the adaptive cross-support window

The adaptive cross-support window is the basis for constructing an adaptive filtering area. Based on the calculation criterion of arm length according to the colour and distance double threshold constraints, we apply a stereo matching algorithm method ((Li and Wang 2020) to generate an adaptive cross-support window and obtain a smooth cross-support window by morphological processing, as shown in Fig. 2. For more efficiency, we optimise the adaptive cross-support window initialisation approach on the CUDA platform.

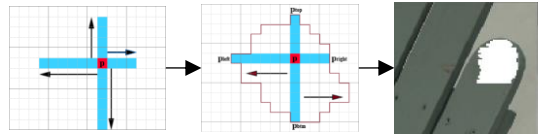

**Fig. 2** Process of generating smooth adaptive cross-support window

4.2 Parallax initialisation

In this stage, we select the initial disparity based on the lowest matching cost to prepare for the next stage of disparity optimisation. This section describes how to calculate the point matching cost to complete the cost initialisation in the adaptive filter area, average the cost for matching cost of the corresponding point in the adaptive filter area, and find the smallest constraint cost as the matching point.

4.2.1 Cost calculation

The cost calculation is the similarity measure of the left and right images pixel by pixel in parallax. We consider fusing the AD and census measures (Zabih and Woodfill 1994) method, called AD-census adaptive measurement (Li and Wang 2020), to construct the adaptive function by exploiting the shortest arm length of the texture-based cross-support window.

The AD-census adaptive measurement function for calculating the matching cost of point $p$ and point $q$ is:

$$C(p,d) = \alpha \left(1 - e^{-\frac{C_{AD}(p, p-d)}{\lambda_{AD}}}\right)$$
$$+ (1-\alpha)\left(1 - e^{-\frac{C_{census}(p, p-d)}{\lambda_{census}}}\right),$$

$$\alpha = 1 - e^{-\frac{\gamma_L}{L_{min}+\varepsilon}}, \tag{1}$$

where $\lambda_{AD}$ and $\lambda_{census}$ are the regularisation parameters for the two basic measures and $p$ is the point to be matched in the left image of the binocular visual image. The pixels with parallax $d$ in the horizontal direction in the right picture are $q = p-d$. Moreover, $\alpha$ is the weight of adjusting the contribution cost of the two measures, which is based on the shortest arm length $L_{min}$, the edge control parameter $\gamma_L$, and the correction parameter $\varepsilon$. In other words, when the shortest arm length becomes longer, the texture of the current region becomes weaker, and the weight smaller. The cost contribution of the AD measurement calculation is reduced, whereas that of the census measurement calculation is increased to achieve the purpose of adaptive cost calculation according to the texture.

### 4.2.2 Cost aggregation

In the cost aggregation stage, the adaptive filtering area is leveraged for cost aggregation. Assuming that pixel $p$ is a pixel in the left view, and $D_p$ is the pixel set in the corresponding adaptive filter area, the matching cost of pixel $p$ within the parallax range $d \in [d_{min}, d_{max}]$ can be finally expressed as

$$C(p,d) = \frac{1}{N}\sum_{p' \in N_p} C(p', d), \tag{2}$$

where $N$ is the number of pixels in the adaptive filter area $N_p$. After calculating the cost of point $p$ under each disparity, a winner-takes-all strategy is adopted, and the smallest matching cost is selected as the initial disparity:

$$d_p = \underset{d \in [d_{min}, d_{max}]}{\operatorname{argmin}} (C(p,d)). \tag{3}$$

### 4.3 Parallax optimisation

In general, there are many mismatched areas in the parallax information provided by the initial parallax, such as occlusion area, image edge distortion, misjudgement of different value points, and noise interference. In our study, the parallax information in each pixel field was used for the statistical optimisation of this point. Our optimisation selects the highest statistical frequency in the neighbour field of pixel $p$, as the parallax value of point $p$, namely:

$$d'_p = \underset{d \in [d_{min}, d_{max}]}{\operatorname{argmax}} (hist(d, N_p)), \tag{4}$$

where $hist(d, N_p)$ is the statistical frequency of parallax $d$ in the neighbourhood $N_p$ of point $p$.

The statistically optimal parallax can be obtained using Eq. (4), which can be applied to perform multiple iterations of parallax to eliminate as many mismatches as possible. Therefore, the outlier points in the parallax map that are clearly in the margin of the parallax value are removed.

### 4.4 Construction of sparse depth map

This subsection describes the conversion of the parallax information into depth information. Because the original image was reduced to a quarter of its original size in the previous calculation of parallax, it is necessary to double all the parallax data before deploying the parameters provided by the binocular module for depth calculation. Fig. 3 depicts some results of sparse depth map construction in a standard dataset (Middlebury 2014).

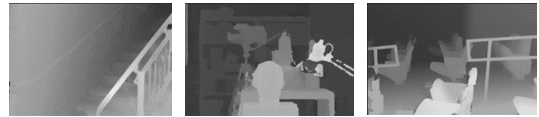

(a) Stairs      (b) Nkuba      (c) Class

**Fig. 3**  Some sparse depth map results of Middlebury 2014 dataset

## 5 Depth contour extraction based on the DIS optical flow method

In the proposed DCO algorithm, the accuracy of occlusion contour depends on the accuracy of depth contour. Therefore, this section will extract the depth contour of the three-dimensional boundary of the object, that is, find the contour edge filter where the object may be occluded, filter the contour information of the whole image, and retain the three-dimensional contour information that may be occluded.

In order to filter the planar texture without three-dimensional features and retain the depth profile on the three-dimensional edge, the depth profile filter needs to be calculated first. In order to obtain the depth profile filter, the gradient amplitude needs to be calculated to extract the occlusion area when the three-dimensional object moves. Then, according to the confidence analysis of optical flow field data, the gradient amplitude region with the highest confidence in the two opposite flow fields is combined and normalized to extract the three-dimensional depth profile. Fig. 4 illustrates the whole process of the DIS-based depth contour filter extraction. The following subsections will introduce the details of bidirectional optical flow field and gradient amplitude calculation methods, bidirectional amplitude fusion and depth contour extraction methods respectively.

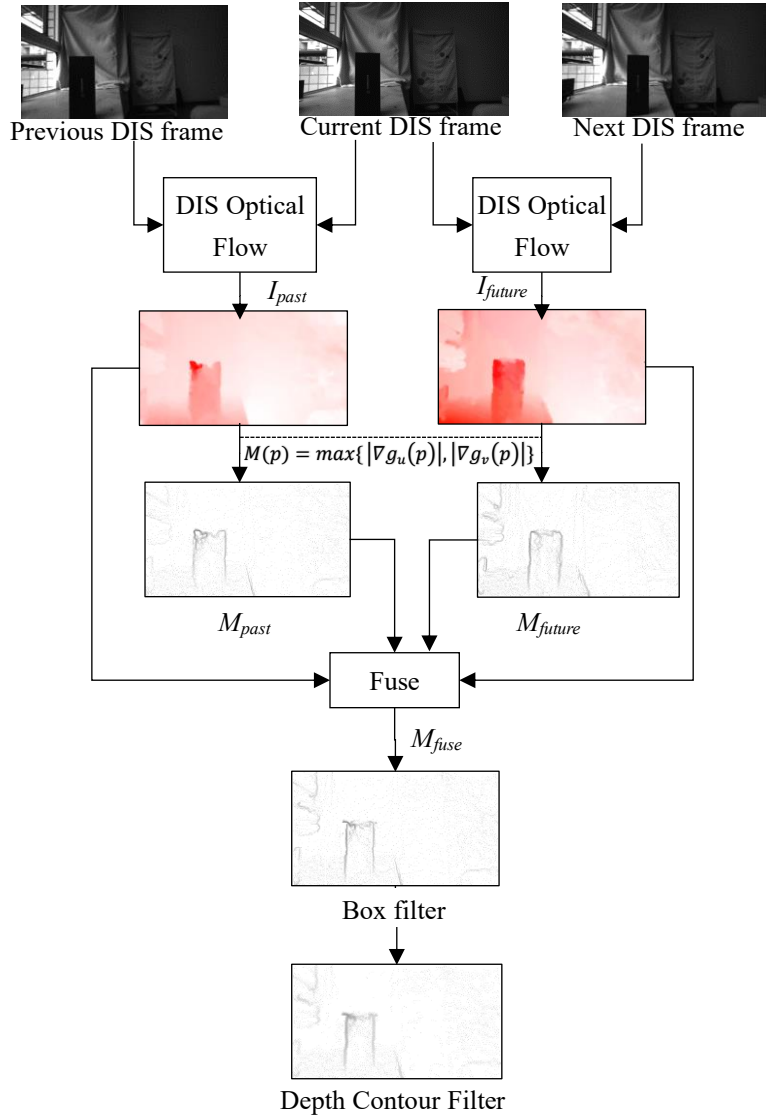

**Fig. 4** Depth contour filter extraction based on DIS optical flow method

5.1 Calculation of bidirectional optical flow field and gradient amplitude

In this step, the optical flow field is used to calculate the gradient amplitude so as to obtain the occlusion area when the three-dimensional object moves. However, the data of moving occlusion area of the unidirectional optical flow field is unreliable. For this reason, it is necessary to maintain an optical flow key frame window with a capacity of three frames, considering the middle frame as the centre; calculate the optical flow field of the adjacent key frames, forward (future frame) and backward (past frame); fuse the reliable amplitude information in two directions.

5.1.1 Bidirectional optical flow field calculation

First, we implement the three frames of the images in the optical flow key frame window to calculate the forward and the backward optical flow field with the DIS optical flow algorithm (Kroeger et al. 2016), obtain $I_{past}$ and $I_{future}$, and convert them into images, as shown in the first row of Fig. 4.

By analysing the preliminary results of the optical flow field calculation, it can be derived that the optical flow data obtained in the occlusion area of the adjacent frame are incorrect. For example, in the forward optical flow field, if the black foreground object in the middle moves relative to the left, the left background of the middle frame is occluded in the previous frame, and the optical flow data of this occluded area are incorrect. Correspondingly, the optical flow data on the right edge of the subject are retained more accurately, and the backward field conditions can be deduced by analogy. For the problem of retaining the most accurate region, a specific solution is provided in the fusion stage in Subsection 5.2.

5.1.2 Gradient amplitude calculation

From the conclusions of the previous subsection, it can be inferred that the places with strong optical flow data changes generally belong to the contour area of the object. Therefore, after obtaining the bidirectional optical flow field, it is necessary to keep the parts with large changes in these data, let:

$$I = \{p \in N_{flow} | U_p, V_p\}. \qquad (5)$$

After obtaining the optical flow field $I$, we convert the optical flow field data to polar coordinates, as shown in Eq. (6):

$$I_{polar} = \{p \in N_{flow} | \theta_p, r_p\}, \qquad (6)$$

where $N_{flow}$ is a plane coordinate space with a horizontal upper limit of $u_{size}$ and a vertical upper limit of $v_{size}$. Because the scalar $r_p$ can reflect the speed of the pixel at point $p$, $r_p$ is used to calculate the motion change rate of adjacent pixels on the x-coordinate $u$ and y-coordinate $v$, as follows:

$$\begin{cases} \nabla g_u(p) = r_{p\_right} - r_p, p\_right \in N_{flow}, \\ \nabla g_v(p) = r_{p_{bottom}} - r_p, p_{bottom} \in N_{flow}. \end{cases} \qquad (7)$$

The amplitude of the pixel motion change rate in the horizontal and vertical directions in the optical flow field can be calculated by Eq. (6) and Eq. (7), and the maximum amount of $r$ change in the two directions is considered as the final gradient amplitude. Hence, it constitutes the complete gradient amplitude matrix, M:

$$M(p) = max\{|\nabla g_u(p)|, |\nabla g_v(p)|\}$$
$$M = \{p \in N_{flow} | M(p)\}. \qquad (8)$$

After visualising the matrix, the effect is illustrated in the second row of Fig. 4. The result indicates that the object part that may have a depth contour is extracted.

5.2 Bidirectional amplitude fusion

To eliminate the unreliable part of the data and retain the reliable one, we propose a fusion gradient amplitude method. It can be observed that the data of $I_{past}$ and $I_{future}$ are exactly complementary in the three input frames. In other words, the occlusion part in one matrix data corresponds to a more reliable part in the other matrix.

To extract reliable data, we started from the optical flow data by establishing a mathematical model and obtaining reliable data from the internal rule. First, observing a certain spatial point in the previous frame, if it is not occluded in the future frame, the optical flow data around the corresponding projected plane coordinates of this part of the space are more reliable. Conversely, if a certain spatial point in the past frame is occluded in the future frame, it is generally unreliable.

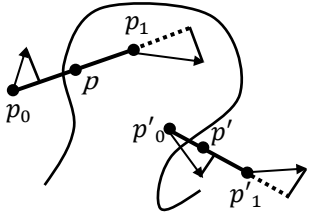

**Fig. 5** Confidence analysis of optical flow data

As depicted in Fig. 5, an object moves to the right in the next frame relative to the previous frame. Now, consider points *p* and *p′*. Point *p* is located near the left edge of the object, and point *p′* is located near the right edge. Consider the two points as the midpoints, and assume the pixel motion direction to be the positive direction to create the 2D vectors $\overrightarrow{p_0p_1}$ and $\overrightarrow{p'_0p'_1}$. Subsequently, find the projection values $f_0$, $f_1$, $f'_0$, and $f'_1$ of the motion vectors corresponding to points $p_0$, $p_1$, $p'_0$, and $p'_1$, respectively:

$$f(p_i) = proj\big(I(p_i), \overrightarrow{pp_i}\big). \tag{9}$$

After expanding and simplifying, we obtain:

$$f(p_i) = I(p_i) \cdot \vec{e}, \tag{10}$$

where $\vec{e}$ is the unit vector corresponding to the optical flow vector of the pixel at point *p*. If the object in Fig. 5 moves to the right and the area where point *p* is located is no longer occluded, the pixel motion intensity at point $p_1$ in the positive direction of the motion vector will be higher than that at point $p_0$ in the reverse direction of the motion vector. Meanwhile, point *p′* is in the occluded area; consequently, the pixel motion intensity at point $p'_1$ in the positive direction of the corresponding motion vector is lower than that at point $p'_0$ in the reverse direction. By observing these, suppose that points *p* and *p′* are located at the same coordinates as $M_{past}$ and $M_{future}$, respectively, and let:

$$\begin{cases} r_{past} = f_1 - f_0, \\ r_{future} = f'_1 - f'_0. \end{cases} \tag{11}$$

Irrespective of the direction in which point *p* moves, its relationship with *r* in the occluded area is constant; hence, the acquisition constraints of each element in the bidirectional amplitude fusion matrix can be obtained as follows:

$$M_{fuse}(p) = \begin{cases} M_{past}(p), r_{past} > r_{future}, \\ M_{future}(p), r_{future} > r_{past}. \end{cases} \tag{12}$$

Essentially, on one hand, the generation of depth contour is followed by the law of complementary motion between the intermediate frame and front-back frames; on the other hand, the generation of depth contour is dependent on the relationship between the area with high confidence in the optical flow data and the occlusion area. The visualisation effect of matrix $M_{fuse}$ after the aforementioned bidirectional amplitude fusion is shown in the third row of Fig. 4 (left).

It can be observed that a depth contour filter with high confidence has been extracted. However, the high-confidence area in the filter is often not large enough to cover the entire depth contour. We use box filtering for $M_{fuse}$ to expand the confidence range to ensure that it covers the entire depth contour filter, as shown in the third row of Fig. 4 (right).

5.3 Depth contour extraction

In our method, we deploy the depth contour filter in the dual threshold detection stage to filter the texture that is not at the 3D contour and retain the depth contour.

5.3.1. Image contour extraction

As the most intuitive and easy-to-extract structural information, image texture edges are the basis for obtaining high-level image information. However, the Canny edges (Canny 1986) obtained from these data alone are rough; for this reason, we performed a non-maximum suppression processing (Neubeck and Gool 2006) to improve the contour extraction.

After non-maximum suppression processing, the processed grey gradient data must be converted into image contours. Here, dual thresholds are exploited to detect and connect the edge contours. First, we set the strong threshold $T_{high}$ and the weak threshold $T_{low}$, which discard the points whose gradient intensity is less than $T_{low}$ and mark the points greater than $T_{high}$ as contours. After coarse filtering, there will be some points between $T_{high}$ and $T_{low}$ in the image. The points in between may be in the contour part or in the non-contour. Here, eight connected regions are exploited to identify those points that should be kept connected to the corresponding pixels higher than $T_{high}$, marking them as edge points, and the rest ones as non-edge points.

### 5.3.2. Non-depth texture contour filtering

To extract the final depth contour, it is necessary to add specified constraints to extract the depth contour information belonging to the edge of the 3D structure.

In Subsection 5.2, $M_{fuse}$ of the input frame by optical flow calculation and bidirectional amplitude fusion is obtained. First, normalise it to make $M_{fuse}(p) \in [0,1]$ and, subsequently, modify the bilateral filtering rules of the original Canny algorithm to add an additional threshold $T_{depth}$. If point $p$ satisfies $M_{fuse}(m(p)) < T_{depth}$, no further processing is performed and the point is directly marked as a non-contour point. The rest of the points within the depth contour range will be further processed by Canny algorithm.

Because the depth contour filter is processed when the size of the original image is reduced, the coordinates of the query point $p$ need to be mapped in the original size contour processing. In this study, the optical flow input image size was scaled down to a quarter of the original image size.

The results show that the depth contour extraction algorithm based on the DIS optical flow method can effectively extract the edge contours of 3D object structures, and filter out the texture of object surfaces that does not contain 3D information, as shown in Fig. 6.

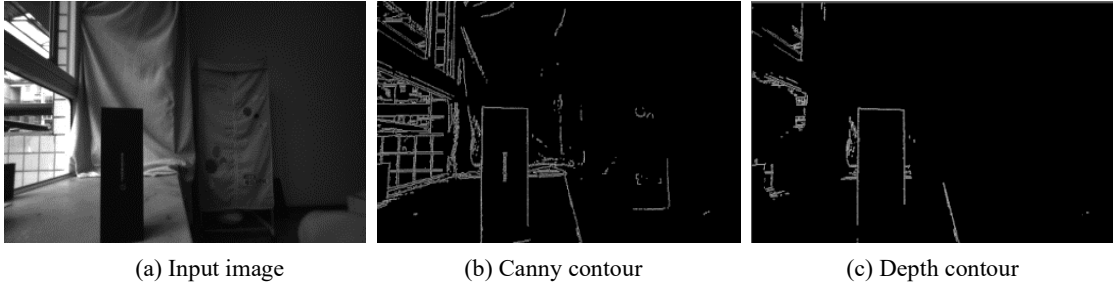

  (a) Input image     (b) Canny contour     (c) Depth contour

**Fig. 6** Results of the contours generated by Canny and the depth contours after texture filtering

## 6 Depth information extraction by quadratic optimisation

We deploy the discrete depth information and depth contour information of the existing sparse depth map and adopted a quadratic optimisation model to infer the depth value on the remaining points without depth information. Furthermore, we incorporate the depth contour of the object into algorithm calculation and calculated the contour-sensitive dense depth map. Finally, we could render the real-virtual occlusion scene.

How to increase the density of sparse depth information can be defined as a quadratic optimisation problem. The constraints proposed by Aleksander et al. (2018) are based on sparse depth map information, SLAM re-projection information, and edge information. In our study, we modify the constraints and refrain from SLAM related constraints. Instead, the constraints area is constructed from the discrete depth information of the image and the depth contour edge information. The following introduce as the construction of the proposed three constraints.

Based on the obtained discrete depth information, the first constraint constructed after the above-mentioned steps is based on the sparse depth map. If the contour-sensitive dense depth map required in the end is $D$, the first constraint is:

$$E_{\text{data}}(p) = w_{\text{sparse}} \left| D(p) - D_{\text{sparse}}(p) \right|^2, \quad (13)$$

where $D_{\text{sparse}}(p)$ is the depth information of the discrete depth map at point *p*, and the value of $w_{sparse}$ satisfies:

$$w_{\text{sparse}} = \begin{cases} 0, D_{\text{sparse}}(p) = 0; \\ 1, D_{\text{sparse}}(p) > 0. \end{cases} \quad (14)$$

In summary, if a point is in the missing information area of the sparse depth map, it has no cost contribution; otherwise, the constraint is valid and included in the cost contribution.

The second constraint needs to consider the discontinuity of the depth information of the depth edge and the smoothness of the depth information of the non-depth edge. This study uses smoothness constraint to set the depth contour information that participates in the optimisation and generates more accurate 3D depth contour information; the smoothness constraint is expressed as follows:

$$E_{\text{smooth}}(p,q) = w_{pq}|D(p) - D(q)|^2. \quad (15)$$

Point *q* is a point in the 4-neighbourhood of point *p*, and the value of $w_{pq}$ satisfies:

$$w_{\text{pq}} = \begin{cases} 0, B_{dp}(p) + B_{dp}(q) = 1 \\ \max(1 - \min(s_q, s_p), 0), else. \end{cases} \quad (16)$$

Normalise the $M_{intensity}$ involved in the Canny contour extraction to the interval [0,1] to obtain the matrix $M_I$. Consequently, $s_q$ and $s_p$ satisfy:

$$\begin{cases} s_p = M_{\text{fuse}}(p) \cdot M_I(p) \\ s_q = M_{\text{fuse}}(q) \cdot M_I(q), \end{cases} \quad (17)$$

where $B_{dp}$ is the binary image of the depth contour. If either point *p* or point *q* is on the depth contour, the depth value at point *p* will not be smoothed; hence, the depth discontinuous area will be aligned with the depth contour. Otherwise, it can be determined that neither of points *p* and *q* is on the depth contour or both points belong to the depth contour. However, in either case, the depth information of the two points is smooth, and if $M_{fuse}$ or $M_I$ is low, indicating that the two points are in the weak texture area or the texture area of the non-depth edge, the value of $w_{pq}$ is larger and the smoothness constraint is stronger.

The third constraint is stability constraint. In the second optimisation, the sparse depth information of the previous frame is combined for calculation to make the propagation of depth information more stable. The stability constraints are as follows:

$$E_{\text{stable}}(p) = w_{\text{stable}}|D(p) - D_{\text{pre}}(p)|^2, \quad (18)$$

where $D_{pre}(p)$ is the dense depth information at point *p* in the sparse depth map of the previous frame, and $w_{stable}$ satisfies:

$$w_{\text{stable}} = \begin{cases} 0, D_{\text{pre}}(p) = 0, \\ 1, D_{\text{pre}}(p) > 0. \end{cases} \quad (19)$$

Similar to the first constraint, if there is no depth information at point *p* in the previous frame, the point has no contribution; otherwise, it provides constraints. After determining the constraints, we can define the quadratic optimisation formula:

$$\underset{D}{argmin} \, (\lambda_d \sum_p E_{\text{data}}(p) +$$
$$\lambda_s \sum_p \sum_{q \in N_4(p)} E_{\text{smooth}}(p,q) +$$
$$\lambda_{s2} \sum_p E_{\text{stable}}(p)), \quad (20)$$

where $\lambda_d$ is depth constraint balance coefficient, $\lambda_s$ is smoothness constraint balance coefficient, and $\lambda_{s2}$ is stability constraint balance coefficient. The proposed three constraints are deployed to adjust the strength of the depth, smoothness, and stability constraints during the quadratic optimisation. The corresponding solution is a set of sparse linear equations that belong to the Poisson problem. Therefore, the LAHBF algorithm (Szeliski 2006) can be implemented to optimise the solution speed of this problem in GPU.

The depth information extraction results are presented in Fig. 7. The results indicate that the quadratic optimisation model can effectively regularise the depth contour of the object in the depth map.

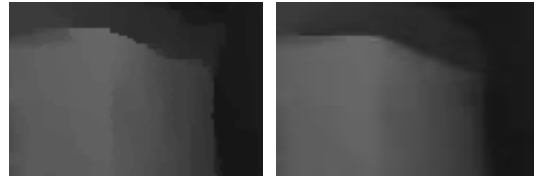

(a) Before optimisation   (b) After optimisation

**Fig. 7** Comparison results of depth information extraction

## 7 Implementation and results

7.1 Implementation details

Our implementation of the DCO algorithm was built on OpenCV, OpenGL, and CUDA platform with the MFC framework. The hardware platform consists of an i5-9600KF CPU, a single GTX1080Ti GPU, an INDEMIND binocular camera as the input device. The experimental input/output frame size is 1280x720. To the best of our knowledge, there is no standard open-source dataset for real-virtual occlusion of binocular frame sequences. Therefore, we collect a small dataset by the easily accessible binocular camera INDEMIND, which is included in supplementary materials.

Fig. 8 demonstrates the implementation process of real-virtual occlusion effect can be obtained in an MR scene, where an OpenGL stencil buffer test is conducted to achieve the real-virtual occlusion effect.

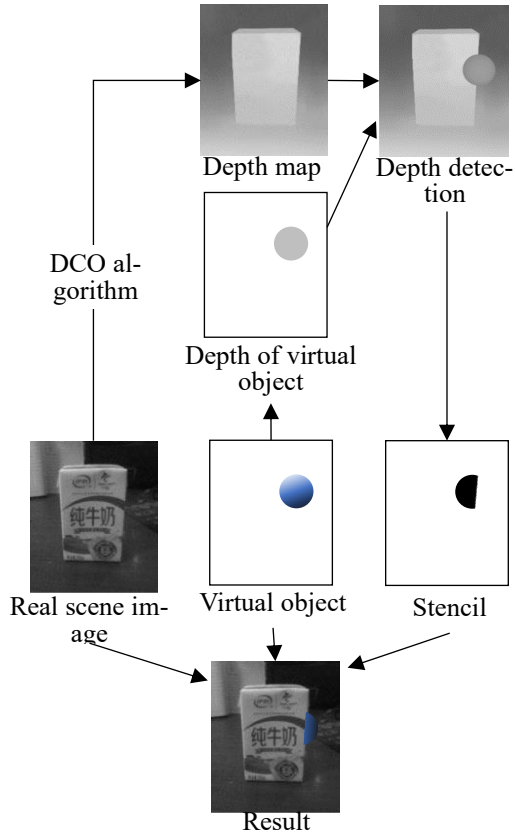

**Fig. 8** Implementation of real-virtual occlusion

7.2 Parameters

**Stereo matching parameters** The values of regularisation parameters $\lambda_{AD}$ is 10.00, and $\lambda_{census}$ is 40.00. Meanwhile, the edge control parameter $\gamma_L$ is set to 1.00, and the correction parameters $\varepsilon$ is set to 0.80.

**Depth contour extraction parameters** For unified processing of image contour extraction the gradient data are normalised to the interval [0,1], $T_{high}$ is set to 0.06, and $T_{low}$ is set to 0.03. Considering that the negative impact caused by the loss of contour information is greater than the influence of contour information redundancy, the threshold $T_{depth}$ in the actual experiment is set to a smaller value of 0.03.

**Depth information extraction parameters** In quadratic optimisation, since the real-virtual occlusion is more accurate than the depth, the smoothing constraint balance coefficient $\lambda_s$ is set to 1.20, the depth constraint balance factor $\lambda_d$ is set to 0.80, and the stability constraint balance factor $\lambda_{s2}$ is set to 0.02.

7.3 Algorithm performance evaluation

To prove the timeliness and effectiveness of the algorithm, we conducted time consumption statistical analysis for each stage of DCO algorithm execution and compared the depth contour algorithm optimised with the algorithm without optimisation.

First, the time consumption experiments are done in each execution stage of the proposed two-stage adaptive AD-census stereo matching algorithm. The running time statistics of randomly selected 10 experiment frames of our dataset are presented in Table 1. In detail, 5 frames (frame 1~5) are selected for testing of dense texture scenes and another 5 frames (frame 6~10) are selected for testing of sparse texture scenes.

From Table 1, it can be noticed that for scenes with sparse textures, the efficiency of this algorithm is lower than that of scenes with dense textures. This happens because the cost aggregation time is directly related to the size of the adaptive filter area. The sparser the texture is, the larger the adaptive filter area will be. Consequently, the construction of the adaptive filter area as well as the matching process will require more time. In general, the efficiency was significantly improved when the CUDA platform is adopted to depth map optimise stage.

**Table 1** Time consumption comparison of stereo matching process (ms)

| Frame No. | Adaptive filter area construction | Initial parallax | Parallax optimisation | Sparse | Total time |
|---|---|---|---|---|---|
| 1 | 3.56 | 15.20 | 6.85 | 0.85 | 26.46 |
| 2 | 3.32 | 14.78 | 7.56 | 0.78 | 26.44 |
| 3 | 4.06 | 15.38 | 7.23 | 0.91 | 27.58 |
| 4 | 3.98 | 15.23 | 6.92 | 0.88 | 27.01 |
| 5 | 4.13 | 16.56 | 7.20 | 0.90 | 28.79 |
| 6 | 7.88 | 18.94 | 6.88 | 0.92 | 34.62 |
| 7 | 7.85 | 17.80 | 6.90 | 0.89 | 33.44 |
| 8 | 8.32 | 18.26 | 7.06 | 0.91 | 34.55 |
| 9 | 8.28 | 19.32 | 6.70 | 0.85 | 35.15 |
| 10 | 7.55 | 19.06 | 7.26 | 1.02 | 34.89 |
| Ave. | 5.89 | 17.05 | 7.06 | 0.89 | 30.89 |

**Table 2** Time consumption of depth contour extraction algorithm (ms)

| Frame No. | Bidirectional optical flow | Optical flow amplitude | Amplitude fusion | Box filter | Normalised | Gaussian filtering | Depth contour extraction | Total time |
|---|---|---|---|---|---|---|---|---|
| 1 | 9.28 | 0.31 | 0.87 | 3.82 | 0.12 | 0.65 | 4.04 | 19.09 |
| 2 | 9.64 | 0.33 | 0.87 | 3.54 | 0.14 | 0.50 | 4.00 | 19.02 |
| 3 | 9.71 | 0.30 | 1.47 | 3.76 | 0.09 | 0.62 | 3.96 | 19.91 |
| 4 | 9.42 | 0.28 | 0.86 | 3.63 | 0.16 | 0.63 | 4.07 | 19.05 |
| 5 | 9.06 | 0.30 | 0.87 | 5.16 | 0.17 | 0.54 | 4.17 | 20.27 |
| 6 | 9.22 | 0.28 | 0.92 | 3.31 | 0.08 | 0.58 | 5.18 | 19.57 |
| 7 | 9.50 | 0.30 | 0.88 | 3.61 | 0.08 | 0.55 | 3.69 | 18.61 |
| 8 | 9.34 | 0.28 | 0.85 | 3.58 | 0.10 | 0.60 | 3.64 | 18.39 |
| 9 | 9.75 | 0.30 | 0.97 | 5.68 | 0.15 | 0.61 | 4.26 | 21.72 |
| 10 | 9.19 | 0.28 | 1.06 | 5.55 | 0.10 | 0.49 | 3.64 | 20.31 |
| Ave. | 9.41 | 0.30 | 0.96 | 4.16 | 0.12 | 0.58 | 4.07 | 19.59 |

Table 2 demonstrates the running time statistics of the algorithms involved in the depth contour extraction stage on the CUDA platform. In the depth contour extraction stage, 10 frames of processed data were randomly selected for experiments. From Table 2, the depth contour extraction algorithm alone with the overhead of data transmission can still maintain efficient calculations.

Next, we integrate the algorithms of each stage into the DCO prototype application for time consumption statistics. The operating environment is the indoor environment, and the records of the average time consumption for each stage are presented in Table 3.

The results indicate that the average frame rate of the prototype is 82.43 ms (12.13 fps), which is close to real-time requirements. If a better GPU is used, it may achieve real-time performance. In addition, the algorithm can be further optimised to reduce its time consumption in the future work discussed in Section 8.

**Table 3** Time consumption of each stage of DCO algorithm (ms)

| Stage | Time consumption |
| --- | --- |
| Algorithm initialisation | 560.85 |
| Sparse depth information | 30.35 |
| Depth contour information | 20.08 |
| Depth densification | 23.25 |
| Rendering | 2.25 |
| Other processing | 6.50 |
| Frame processing | 82.43 |

7.4 Algorithm effect evaluation

To verify the reliability of the real-virtual occlusion effect, we conduct an experimental analysis on single-sided, enclosed, and complex occlusions and compare it with the occlusion method of the two-stage adaptive AD-census stereo matching algorithm without depth contour optimisation.

First, we conduct an experimental test for single-sided occlusion to verify whether the depth contour optimisation method in this study applied improved the accuracy of the occlusion contour edge in general. Single-sided occlusion means that the virtual object is completely behind the real object. The virtual object deployed in the experiment was a cube to simulate the comparison result of single-sided occlusion by human hands. The result is demonstrated in Fig. 9.

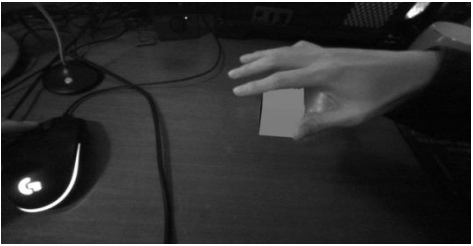
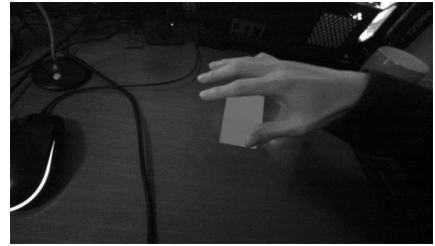

(a) AD-census only method  (b) DCO method

**Fig. 9** Comparison results of single-sided occlusion

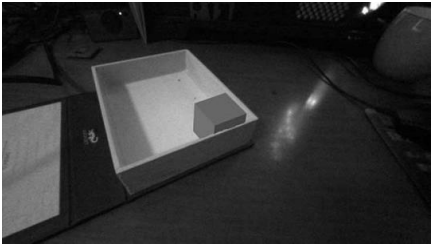
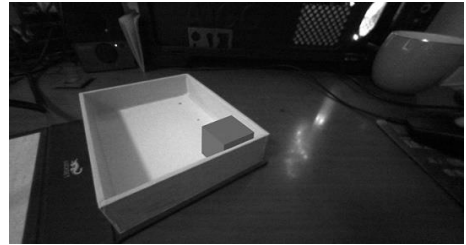

(a) AD-census only method  (b) DCO method

**Fig. 10** Comparison results of the enclosed structures occlusion

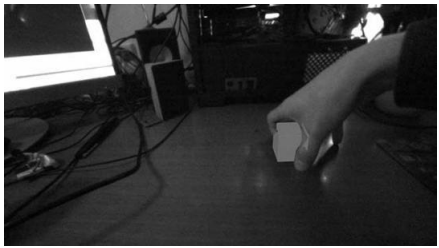
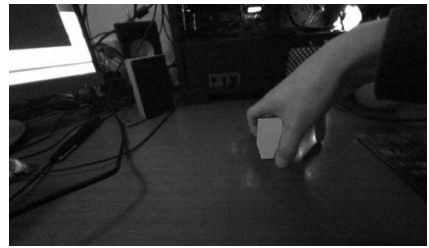

(a) AD-census only method  (b) DCO method

**Fig. 11** Comparison results of complex occlusion

Fig. 9 (a) depicts the result of the disabled depth contour extraction and quadratic optimisation. Fig. 9 (b) illustrates the result of implementing the complete DCO algorithm. By analysing them, it can be deducted that the former is clearly more accurate and reliable than the latter; furthermore, there are still some false occlusions in the un-occluded part of the latter's cube. The results imply that the DCO algorithm has a reliable performance in the accuracy of occlusion contours.

Next, we use objects with enclosed structures to test. The comparison algorithm is the real-virtual occlusion algorithm between the DCO real-virtual occlusion algorithm deployed in this topic and the two-stage adaptive AD-census stereo matching algorithm without depth contour optimisation. The comparison results are presented in Fig. 10.

As depicted in Fig. 10, this group of experiments use boxes to test the depth accuracy of real-virtual occlusion and smooth occlusion accuracy. Both depth judgements are relatively accurate, and there are no abnormalities, such as mould penetration or voids. Some DCO algorithms with apparent contours and occlusion algorithms that do not apply depth contour optimisation have better contour edges. However, the former has greater accuracy than the latter, and the edges are very straight, whereas the latter retains the phenomenon that the occlusion contour is not aligned with the edge of the real object.

Finally, we perform a test to verify the effectiveness and robustness of the DCO algorithm in complex occlusion. Particularly, we use a grasping type to 'grasp' a virtual object. The test results are demonstrated in Fig. 11.

After experimental comparison, it can be inferred that the DCO real-virtual occlusion can additionally obtain the correct occlusion results in more complicated occlusion situations. However, the real-virtual occlusion that deploys solely two-stage adaptive AD-census stereo matching has not only inaccurate contours but also false occlusions. This is because in AD-census only method, the outlier points are not eliminated in depth map construction. Thereby, compared to the proposed DCO method, the depth data generated by AD-census only method are abnormal, which causes abnormal occlusion results.

## 8 Conclusion

In this paper, we proposed a DCO algorithm to handle real-virtual occlusion. The proposed method was based on the sensitivity of contour occlusion and a binocular stereoscopic vision device. First, we reduced the size of the input images to a quarter of the original ones and implemented a two-stage adaptive filter area stereo matching algorithm, AD-census stereo matching, to establish a sparse depth map. The purpose of that stage was to improve the efficiency and obtain more accurate depth information. Second, the DIS optical flow method was used to extract the depth contour of the real object. Third, we proposed three constraints to adjust the strength of the depth, smoothness, and stability constraints during the quadratic optimisation to propagate the sparse depth map to dense depth map. Finally, we implemented a DCO algorithm and evaluated the effectiveness and reliability of the real-virtual occlusion effect. Through experimental comparisons, we proved that our method had satisfactory stability, real-time performance, and effectiveness. Taken together, these results suggest that the DCO algorithm could be further tested and developed in more MR applications.

Future work will overcome the limitations of the proposed method in the following ways: 1) we may combine the SLAM algorithm and depth-point re-projection to build a more efficient computing model; 2) artificial intelligence algorithms can be leverage to optimise depth information extraction and depth contour extraction, with deep neural network model applied. Since the deep learning method need to train/test on a dataset of a certain amount, more data should be collected by the binocular stereoscopic vision device first.

**Acknowledgements** This study is supported by Key Research and Development Plan of Zhejiang Province, China (No.2019C03131), the Public Welfare Technology Application Research Project of Zhejiang Province, China (No. LGF21F020004), Key Lab of Film and TV Media Technology of Zhejiang Province, China (No.2020E10015).